\documentclass[journal]{IEEEtran}
\usepackage{amsmath,amsfonts}
\usepackage{algorithmic}
\usepackage{algorithm}
\usepackage{array}
\usepackage{textcomp}
\usepackage{stfloats}
\usepackage{url}
\usepackage{verbatim}
\usepackage{graphicx}
\usepackage{cite}
\graphicspath{{Figures/}}
\usepackage[caption=false,font=footnotesize]{subfig}

\usepackage{amsthm, amssymb}
\usepackage{mathabx}

\begin{document}

\title{Inter-Frame Compression for Dynamic Point Cloud Geometry Coding}

\author{Anique~Akhtar,~\IEEEmembership{Member,~IEEE,}
        Zhu~Li,~\IEEEmembership{Senior~Member,~IEEE,}\\
        Geert Van der Auwera,~\IEEEmembership{Senior~Member,~IEEE,}
\thanks{This work is supported in part by NSF awards 1747751 and 2148382.}
\thanks{A. Akhtar and G. Van der Auwera is with Qualcomm Technologies Inc., San Diego, CA, 92121, USA (e-mail: aniquea@qti.qualcomm.com, geertv@qti.qualcomm.com).}
\thanks{Z. Li is with the Department of Computer Science and Electrical Engineering, 
University of Missouri-Kansas City, Kansas City, MO 64110 USA (e-mail: zhu.li@ieee.org).}
}


\maketitle

\begin{abstract}
Efficient point cloud compression is essential for applications like virtual and mixed reality, autonomous driving, and cultural heritage. This paper proposes a deep learning-based inter-frame encoding scheme for dynamic point cloud geometry compression. We propose a lossy geometry compression scheme that predicts the latent representation of the current frame using the previous frame by employing a novel feature space inter-prediction network. The proposed network utilizes sparse convolutions with hierarchical multiscale 3D feature learning to encode the current frame using the previous frame. 
The proposed method introduces a novel predictor network for motion compensation in the feature domain to map the latent representation of the previous frame to the coordinates of the current frame to predict the current frame's feature embedding. 
The framework transmits the residual of the predicted features and the actual features by compressing them using a learned probabilistic factorized entropy model. At the receiver, the decoder hierarchically reconstructs the current frame by progressively rescaling the feature embedding. The proposed framework is compared to the state-of-the-art Video-based Point Cloud Compression (V-PCC) and Geometry-based Point Cloud Compression (G-PCC) schemes standardized by the Moving Picture Experts Group (MPEG). The proposed method achieves more than $88\%$ BD-Rate (Bj{\o}ntegaard Delta Rate) reduction against G-PCCv20 Octree, more than $56\%$ BD-Rate savings against G-PCCv20 Trisoup, more than $62\%$ BD-Rate reduction against V-PCC intra-frame encoding mode, and more than $52\%$ BD-Rate savings against V-PCC P-frame-based inter-frame encoding mode using HEVC. These significant performance gains are cross-checked and verified in the MPEG working group.
\end{abstract}


\section{Introduction}
A point cloud (PC) is a 3D data representation that is essential for tasks like virtual reality (VR) and mixed reality (MR), autonomous driving, cultural heritage, etc. PCs are a set of points in 3D space, represented by their 3D coordinates (\textit{x, y, z}) referred to as the \textit{geometry}. Each point may also be associated with multiple \textit{attributes} such as color, normal vectors, and reflectance. Depending on the target application and the PC acquisition methods, the PC can be categorized into point cloud scenes and point cloud objects. Point cloud scenes are typically captured using LiDAR sensors and are often dynamically acquired. Point cloud objects can be further subdivided into static point clouds and dynamic point clouds. A static PC is a single object, whereas a dynamic PC is a time-varying PC where each instance of a dynamic PC is a static PC. Dynamic time-varying PCs are used in AR/VR, volumetric video streaming, and telepresence and can be generated using 3D models, i.e., CGI, or captured from real-world scenarios using various methods such as multiple cameras with depth sensors surrounding the object. These PCs are dense photo-realistic point clouds that can have a massive amount of points, especially in high precision or large-scale captures (millions of points per frame with up to 60 frames per second (FPS)). Therefore, efficient point cloud compression (PCC) is particularly important to enable practical usage in VR and MR applications.
This paper focuses on geometry compression for the dense dynamic point clouds. Temporally successive point cloud frames share some similarities, motion estimation is key to effective compression of these sequences. However, these frames may have different numbers of points, and exhibit no explicit association between points over time. Performing motion estimation, motion compensation, and effective compression of such data is, therefore, a challenging task.

The Moving Picture Experts Group (MPEG) has approved two PCC standards \cite{emerging, gpcc_vpcc}: Geometry-based Point Cloud Compression (G-PCC) \cite{gpcc} and Video-based Point Cloud Compression (V-PCC) \cite{vpcc}. 
G-PCC includes octree-geometry coding as a generic geometry coding tool and a predictive geometry coding (tree-based) tool which is more targeted toward LiDAR-based point clouds. G-PCC is still developing a triangle meshes or triangle soup (trisoup) based method to approximate the surface of the 3D model. V-PCC on the other hand encodes dynamic point clouds by projecting 3D points onto a 2D plane and then uses video codecs, e.g., High-Efficiency Video Coding (HEVC), to encode each frame over time. MPEG has also proposed common test conditions (CTC) to evaluate test models \cite{CTC}.

Deep learning solutions for image and video encoding have been widely successful \cite{liu2020deep}. Recently, similar deep learning-based PC geometry compression methods \cite{quach2019learning, wang2021lossy, adl, GeoCNNv2, Muscle, octsqueeze, voxelcontext, octattention, gao2021point, you2021patch, PCGCv2, wang2021sparse} have been shown to provide significant coding gains over traditional methodologies. Point cloud compression represents new challenges due to the unique characteristics of PC. For instance, the unstructured representation of PC data, the sparse nature of the data, as well as the massive number of points per PC, specifically for dense photo-realistic PC, makes it difficult to exploit spatial and temporal correlation. The current deep learning-based PC geometry compression solutions are all intra-prediction methods for static point clouds and fail to utilize inter-prediction coding gains by predicting the current frame using previously decoded frames. 

Inter-prediction schemes in video compression are very successful in performing motion compensation to achieve impressive results. However, similar motion compensation for dynamic point clouds is not possible because the coordinates between different frames of a point cloud sequence are different due to non-uniform sampling in the spatial-temporal space of the point cloud geometry. Performing motion estimation across frames with changing voxels occupancy is challenging and hence the deep-learning solutions struggle to perform motion compensation on dynamic point cloud frames. To this end, we propose a novel inter-frame point cloud compression scheme that successfully performs motion compensation. Following MPEG's PCC category guidelines, our work seeks to target dense dynamic point clouds used for VR/MR and immersive telecommunications. Sparse dynamically acquired LiDAR-based point clouds are a very different point cloud category that is out of the scope of this work.
Our contributions are summarized as follows:
\begin{itemize}
  \setlength\itemsep{0em}
  \item A novel deep learning-based framework is proposed for point cloud geometry inter-frame encoding similar to P-frame encoding in video compression.
  \item We propose a novel inter-prediction module (predictor network) that learns a feature embedding of the current PC frame from the previous PC frame. The network utilizes hierarchical multiscale feature extractions and employs a generalized sparse convolution (\textit{GSConv}) with arbitrary input and output coordinates to perform motion compensation in the feature domain by mapping the latent features from the coordinates of the first frame to the coordinates of the second frame. The inter-prediction module is the first deep learning module that successfully enables the effective transferring of features between point cloud frames with different coordinates.
\end{itemize}
Experimental results show the proposed method achieving more than $88\%$ BD-Rate gains against G-PCCv20 (octree), more than $56\%$ BD-Rate gains against G-PCC (trisoup), more than $34\%$ BD-Rate gains against state-of-the-art deep learning-based point cloud geometry compression method, more than $62\%$ BD-Rate gains against V-PCCv18 intra-frame mode, and more than $52\%$ BD-Rate gains against V-PCCv18 P-frame-based (low-delay) inter-frame mode which uses HEVC.

\section{Background}
Our research is most closely related to three research topics: point cloud geometry compression, deep learning-based video inter-frame coding, and deep learning-based point cloud compression. 

Prior non-deep learning-based point cloud geometry compression mostly includes \textit{octree-based, triangle mesh-based}, and \textit{3D-to-2D projection-based} methodologies. \textbf{Octree-based methods} are the most widely used point cloud encoding methods \cite{octree0, octree1, octree2}. Octree provides an efficient way to partition the 3D space to represent point clouds and is especially suitable for lossless coding. In these methods, the volumetric point cloud is recursively divided into octree decomposition until it reaches the leaf nodes. Then the occupancy of these nodes can be compressed through an entropy context modeling conditioned on neighboring and parent nodes. Thanou et al. \cite{graph} implemented octree-based encoding for time-varying point clouds that can predict graph-encoded octree structures between adjacent frames. 
MPEG's G-PCC standard \cite{emerging} also employs an octree-based compression method known as \textit{octree geometry codec} and is specifically devoted to sparse point clouds. G-PCC encoding can further be complemented by triangle meshes (a.k.a., triangle soups) which are locally generated together with the octree to terminate the octree decomposition prematurely. This helps reconstruct object surfaces with finer spatial details and is known as the \textit{trisoup geometry codec} \cite{gpcc_vpcc2}.

\textbf{3D-to-2D projection-based methods}. Traditional 2D image and video coding have demonstrated outstanding efficiency and have been widely used in standards which have motivated works to project 3D objects to multiple 2D planes and leverage popular image and video codecs for compact representation. MPEG's V-PCC \cite{gpcc_vpcc} standard is one such 3D-to-2D projection-based solution that is specifically designed for dense, as well as, dynamic PCs. The V-PCC  standard projects the points and the corresponding attributes onto planes and then uses a state-of-the-art video codec, such as HEVC, to encode point clouds. V-PCC has both intra-frame coding as well as inter-frame coding \cite{li2019advanced} where the previously decoded frames are employed to encode the next frames. We have recently also had some works for \textbf{dynamic point cloud compression} \cite{dynamic_deep_1, dynamic_deep_2, dynamic_deep_3}. However, their results are still lacking and the performance is not comparable to V-PCC.

\textbf{Deep learning-based models for image and video encoding} can learn an optimal non-linear transform from data along with the probabilities required for entropy coding the latent representation into a bitstream in an end-to-end fashion. For image compression, autoencoders \cite{balle2016end} were initially adopted and the best results were achieved by employing variational autoencoders with side information transmission and applying an autoregressive model \cite{balle2018variational}. Deep learning solutions for video compression methods usually employ 3D autoencoders, frame interpolation, and/or motion compensation via optical flow. 3D autoencoders are an extension of deep learned image compression. Frame interpolation methods use neural networks to temporally interpolate between frames in a video and then encode the residuals \cite{video1}. Motion compensation via optical flow is based on estimating and compressing optical flow which is applied with bilinear warping to a previously decoded frame to obtain a prediction of the frame currently being encoded \cite{video2}. Current deep learning-based PCC takes inspiration from the deep learning-based image compression methods but so far has not been able to implement inter-frame prediction models commonly used in video encoding. Our work is the first method that takes inspiration from the frame interpolation-based methods in video encoding to perform inter-frame encoding for dynamic point clouds.

Deep learning-based Geometry PCC can be broadly categorized into: \textit{voxelization-based methods, octree-based methods, point-based methods}, and \textit{sparse tensors-based methods}. \textbf{Voxelization-based methods} were employed in the earlier approaches, including Quach et al. \cite{quach2019learning}, Wang et al. \cite{wang2021lossy}, Guarda et al. \cite{adl} and Quach et al. \cite{GeoCNNv2}. These methods voxelize the PC and then divide it into smaller blocks typically of $64\times64\times64$ voxels. Then 3D convolutions are applied using autoencoder architectures to compress these blocks into latent representations. These methods usually employ a focal loss or a weighted binary cross-entropy loss to train their model. However, these methods also have to process empty voxels which are usually the majority of the voxels and are, therefore, computational and memory inefficient. 

\begin{figure*}[!t]
\centering
\includegraphics[width=\linewidth]{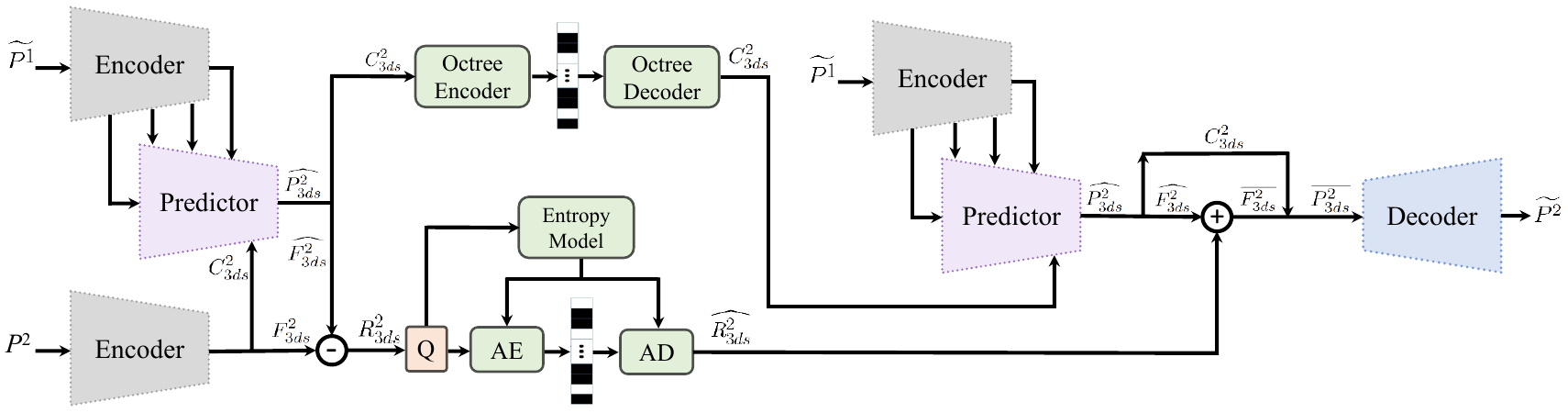}
\caption{System Model. The previously decoded frame $\widetilde{P^1}$ is employed to encode a feature embedding of the current frame $P^2$. Multiscale features from $\widetilde{P^1}$ and three-times downsampled coordinates $C^2_{3ds}$ from $P^2$ are passed to the Predictor network to learn a feature embedding $\widehat{P^2_{3ds}} = \{C^2_{3ds}, \widehat{F^2_{3ds}}\}$. The current frame's three-times downsampled coordinates $C^2_{3ds}$ are transmitted in a lossless manner using an octree encoder. The predicted downsampled features $\widehat{F^2_{3ds}}$ and the original downsampled features $F^2_{3ds}$ are subtracted to obtain the residual features $R^2_{3ds}$. The residual is transmitted in a lossy manner using a learned entropy model. The same Encoder and Predictor module are used throughout the system. Q, AE, and AD stand for quantization, arithmetic encoder, and arithmetic decoder respectively.}
\label{system}
\vspace{-2mm}
\end{figure*}

\textbf{Octree-based deep learning methods} employ octree representation to encode the PCs leading to better consumption of storage and computation. These methods employ entropy context modeling to predict each node's occupancy probability conditioned on its neighboring and parent nodes. MuSCLE \cite{Muscle} and OctSqueeze \cite{octsqueeze} employ Multi-Layer Perceptrons (MLPs) to exploit the dependency between parent and child nodes. VoxelContext-Net \cite{voxelcontext} employs both neighbors and parents as well as voxelized neighborhood points as context for probability approximation. Recently, OctAttention \cite{octattention} has been introduced that increases the receptive field of the context model by employing a large-scale transformer-based context attention module to estimate the probability of occupancy code. All of these methods encode the point cloud in a lossless manner and show promising results, particularly on sparse LiDAR-based point clouds.

\textbf{Point-based methods} directly process raw point cloud data without changing their representation or voxelizing them. They typically employ PointNet \cite{pointnet} or PointNet++ \cite{pointnetplus} type architectures that process raw point clouds using point-wise fully connected layers. These methods are typically patch-based methods that employ farthest point sampling to subsample and a knn search to find per point feature embedding to build an MLP-based autoencoder. However as seen in some of these works \cite{gao2021point, you2021patch, huang20193d}, the coding efficiency of such point-wise models is still relatively low and fails to generalize to large-scale dense point clouds. Furthermore, these methods require a lot of pre and post-processing making the encoding process computationally inefficient.

Recent \textbf{sparse convolution-based methods} \cite{PCGCv2, wang2021sparse, losslessSparseContext} have shown really good results especially for denser photo-realistic point clouds. Sparse convolutions exploit the inherent sparsity of point cloud data for complexity reduction allowing for very large point clouds to be processed by a deeper sparse convolutional network. This allows the network to better capture the characteristics of sparse and unstructured points and better extraction of local and global 3D geometric features. However, all of these works employ intra-frame encoding for static point clouds. We employ sparse convolution-based autoencoder architecture similar to \cite{PCGCv2} and design a sparse convolutional inter-frame prediction module that encodes the next PC frame using the previously decoded PC frame similar to P-frame prediction in video encoding.

\begin{figure*}[!t]
\centering
\includegraphics[width=0.75\linewidth]{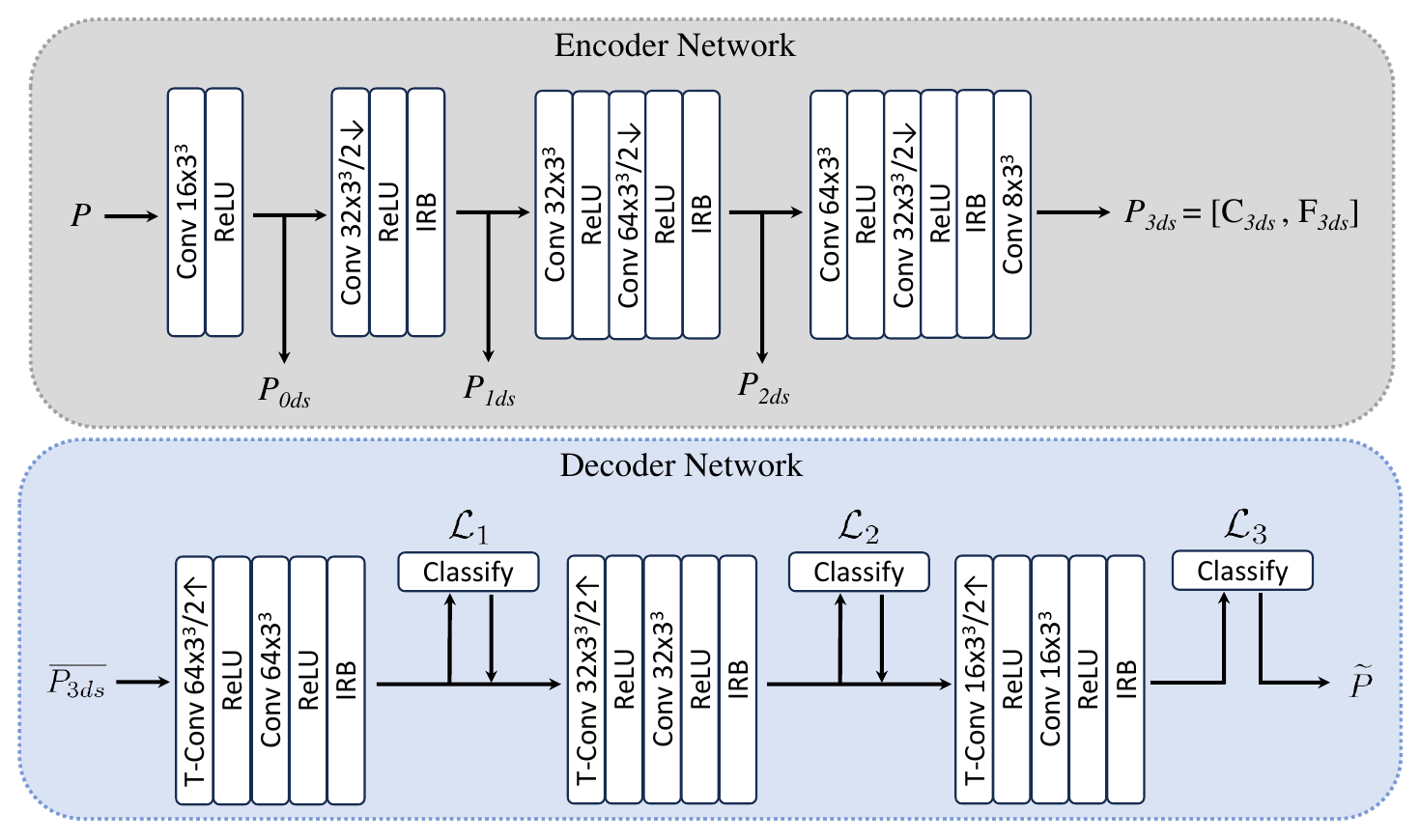}
\caption{Encoder and Decoder Network. The encoder network takes the original point cloud sparse tensor $P$, and creates sparse features at four different scales: $P_{0ds}$, $P_{1ds}$, $P_{2ds}$, and $P_{3ds}$. Where $P_{3ds}$ denotes three-times downsampled sparse tensor containing both the coordinates $C_{3ds}$ and their respective features $F_{3ds}$. The decoder network takes the three-times downsampled sparse tensor and hierarchically reconstructs the original point cloud by progressively rescaling. The decoder upsamples the sparse tensor one scale at a time using transpose convolution followed by classification and pruning to prune out the false voxels.}
\label{encoder_decoder}
\vspace{-1mm}
\end{figure*}

\section{Proposed Method}
\label{sec:method}
The proposed lossy inter-frame point cloud geometry compression framework is illustrated in Fig. \ref{system}. We employ sparse tensors and sparse convolutions to decrease the computational complexity of the network so it can process two PC frames. The solution takes inspiration from the PCGCv2 \cite{PCGCv2} multiscale point cloud geometry compression (PCGC) work. PCGCv2 is an intra-frame point cloud compression scheme suitable for static point clouds. 
The proposed inter-frame compression framework employs an encoder and decoder network similar to PCGCv2 along with a novel inter-prediction module to predict the feature embedding of the current PC frame from the previous PC frame. 
The proposed inter-prediction module employs a specific version of generalized sparse convolution \cite{minkowski} with different input and output coordinates denoted as \textit{GSConv} to perform motion estimation in the feature domain. 
The inter-prediction module is a standalone module that can be employed with different network architectures.
The residual between the predicted and ground truth features are calculated and then these residuals along with the three-times downsampled coordinates are transmitted to the receiver. The three-times downsampled coordinates are losslessly encoded by an octree encoder using G-PCC \cite{gpcc}, whereas the residual features are encoded in a lossy manner using factorized entropy model to predict the probability distribution for arithmetic coding. It should be noted that in our system, the encoder and prediction network is present both at the transmitter as well as the receiver. We train the networks with joint reconstruction and bit-rate loss to optimize rate distortion.
We provide a detailed description of all our modules in subsequent discussions.

\subsection{Problem Formulation and Preprocessing}
We adopt sparse convolutions for low-complexity tensor processing and build our system using Minkowski Engine \cite{minkowski}. Each point cloud frame is converted into a sparse tensor $P$. Each point cloud tensor $P = \{C_n, F_n\}_n$ is represented by a set of coordinates $C = \{(x_n,y_n,z_n)\}_n$ and their associated features $F = \{f(x_n,y_n,z_n)\}_n$. Only the occupied coordinates are kept in a sparse tensor. To initialize the input point cloud as geometry only, we assign feature $f(x,y,z) = 1$ to each occupied coordinate. 
Given a dynamic point cloud with multiple frames, $P^i$, our goal is to convert them into a latent representation with the smallest possible bitrate. We use P-frame encoding where the current frame is encoded using the prediction from the previous frame. We denote the Encoder network as $E$, and the Decoder network as $D$.

\begin{figure}[!t]
\centering
\includegraphics[width=0.8\linewidth]{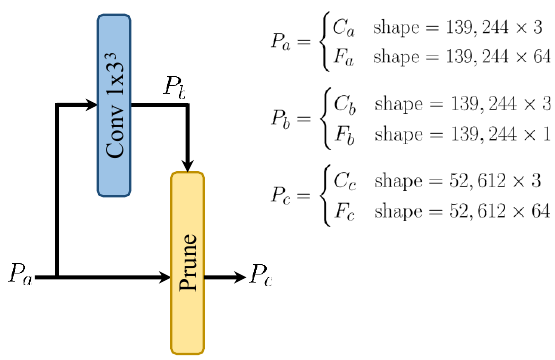}
\caption{Example of classification and pruning layer with input sparse tensor $P_a$ and output sparse tensor $P_c$. Binary classification is applied to $P_b$ to chose the top voxels and prune false voxels from $P_a$ to obtain $P_c$.}
\label{pruning}
\vspace{-1mm}
\end{figure}

\subsection{Feature Extraction}
Our encoder and decoder network is shown in Fig. \ref{encoder_decoder}. We utilize the Inception-Residual Block (IRB) \cite{inception} for feature extraction in all our networks. Each IRB contains three Inception-Residual Network (IRN) similar to PCGCv2 \cite{PCGCv2}.
We employ a multiscale re-sampling with downscaling at the encoder and upscaling at the decoder. This helps exploit the sparsity of the PC while encoding 3D geometric structural variations into feature attributes of the latent representation. 
The encoder is used as a feature extraction module to obtain PC tensors at four different scales capturing multiscale features at different level of details: $P_{0ds}, P_{1ds}, P_{2ds}, P_{3ds} = E(P)$. 
Where $P_{jds}$ represents a sparse tensor P that has been downsampled $j$ times.

\subsection{Point Cloud Reconstruction}
The decoder receives a three-times downsampled PC tensor and upsamples it hierarchically to reconstruct the original PC tensor by employing a different reconstruction loss at each scale: $\widetilde{P} = D(\widebar{P_{3ds}})$. Decoder employs transpose convolution to upsample the PC tensor and generate newer voxels. After each upsampling, the probability of voxel occupancy $p_v$ is predicted and a binary classification loss is employed. The geometry at the decoder is reconstructed by employing a classification and pruning layer to prune false voxels and extract true occupied voxels using binary classification after each upscaling. 
We employ binary cross-entropy loss for voxel occupancy classification as the distortion loss in each pruning layer at the decoder:
\begin{equation}
\mathcal{L}_{BCE} = \frac{1}{N} \sum_{v} -(\mathcal{O}_v \log p_v + (1 - \mathcal{O}_v) \log(1 - p_v))
\end{equation}
where $O_v$ is the ground truth of whether the voxel $v$ is occupied (1) or unoccupied (0).

One example of a classification and pruning layer is shown in Fig. \ref{pruning}. In this example, the input sparse tensor $P_a$ has coordinates $C_a$ of shape $139{,}244\times3$ and their corresponding features of shape $139{,}244\times64$. We pass $P_a$ through a convolution of channel size $1$ to obtain sparse tensor $P_b$ with features $F_b$ of shape $139{,}244\times1$. From $F_b$ we select the top k features (in this example $k=52{,}612$) with the highest probability of occupancy ($p_v$) and their corresponding coordinates using binary voxel classification. 
The false coordinates and their corresponding features are then pruned from $P_a$ to obtain $P_c$. The binary cross entropy loss is employed at each of the pruning layers and is applied to tensor $P_b$ so the true voxels could have a higher $p_v$ and the false voxels have a lower $p_v$. Then the top $k$ voxels with the highest probability of occupancy ($p_v$) are chosen and the rest of the false voxels are pruned out.
$k$ is a metadata calculated at the encoder and losslessly transmitted to the receiver with a very small overhead along with other overhead bits. $k$ determines the number of occupied voxels at each scale for that particular frame which is employed during point cloud reconstruction at the decoder.

\begin{figure*}[!t]
\centering
\includegraphics[width=0.75\linewidth]{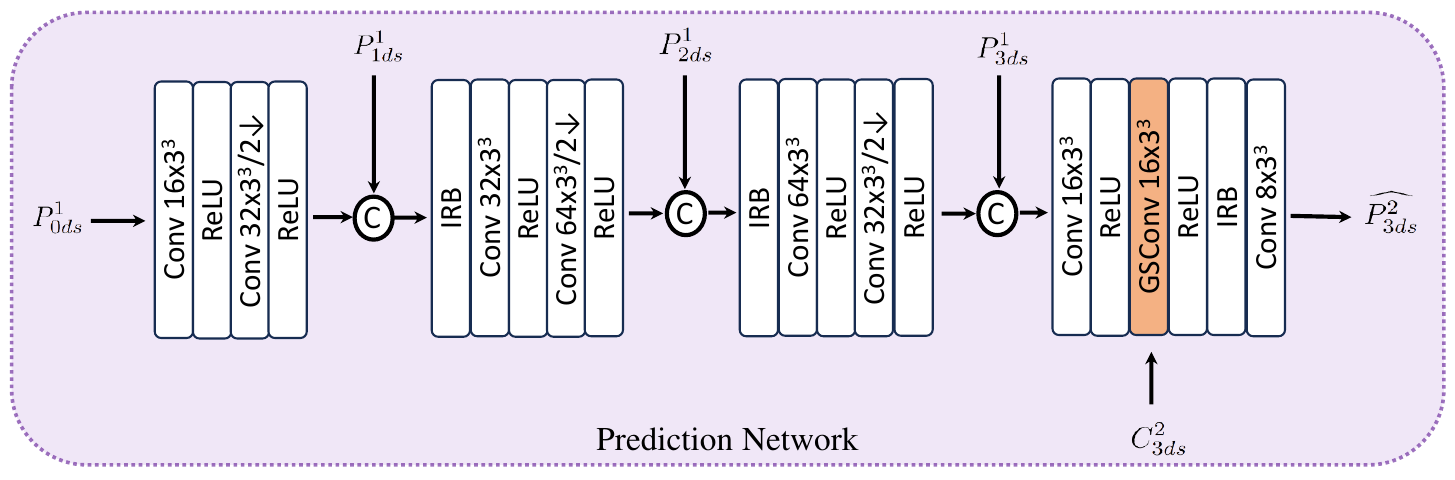}
\caption{Prediction network. Takes in four multiscale features from the previous frame and the three-times downsampled coordinates of the current frame $(C^2_{3ds})$ to learn the current frame's feature embedding $\widehat{P^2_{3ds}}$.}
\label{predictor}
\vspace{-1mm}
\end{figure*}

\subsection{Overall System Model}
This subsection intends to explain the working of the overall system framework shown in Fig. \ref{system}. In our work, we denote the current PC frame as $P^2$ while the previously decoded PC frame is denoted by $\widetilde{P^1}$. The same encoder and prediction module are used throughout the system to decrease the number of parameters. Previously decoded frame $\widetilde{P^1}$ is passed through the encoder to obtain multiscale tensors $P^1_{0ds}, P^1_{1ds}, P^1_{2ds}, P^1_{3ds}$. The current frame ($P^2$) is also passed through the encoder to obtain three-times downsampled tensor containing coordinates and features: $P^2_{3ds} = \{C^2_{3ds}, F^2_{3ds}\}$. Current frame's three-times downsampled coordinates ($C^2_{3ds}$) and the multiscale features from the previous frame are passed to the prediction network to obtain current frame's predicted three-times downsampled tensor $\widehat{P^2_{3ds}} = \{C^2_{3ds}, \widehat{F^2_{3ds}}\}$. The predicted downsampled features $\widehat{F^2_{3ds}}$ and the original downsampled features $F^2_{3ds}$ are subtracted to obtain the residual features $R^2_{3ds}$. The residual is transmitted in a lossy manner using a factorized entropy model \cite{balle2016end}. The current frame's three-times downsampled coordinates $C^2_{3ds}$ are transmitted in a lossless manner using an octree encoder like G-PCC \cite{gpcc}. Three-times downsampled coordinates $C^2_{3ds}$ is much smaller than the original geometry (e.g. for the 8iVFB dataset \cite{8ijpeg}, the $C^2_{3ds}$ is about $16$ times smaller than $C^2$). At the receiver, the previously decoded frame $\widetilde{P^1}$ and the three-times downsampled coordinates $C^2_{3ds}$ are used to predict $\widehat{P^2_{3ds}}$. The residual $\widehat{R^2_{3ds}}$ is added with $\widehat{P^2_{3ds}}$ to obtain the current frame's three-times downsampled tensor representation $\widebar{P^2_{3ds}}$. The decoder progressively rescales $\widebar{P^2_{3ds}}$ to obtain the current decoded frame $\widetilde{P^2}$. Encoder and decoder architecture can on their own be used without the prediction module for intra-frame PC compression.

\subsection{Inter-Prediction Module}
Until now, efficient motion estimation for dynamic point clouds has not been possible due to the difference in the occupied coordinates between point cloud frames. 
We propose a novel deep learning-based inter-frame predictor network that can predict the latent representation of the current frame from the previously reconstructed frame as shown in Fig. \ref{predictor}. 
This is the first inter-prediction module that is capable of performing motion estimation of 3D coordinates across frames such as to learn the current frame's feature embedding. The framework does not explicitly performs motion estimation of distinct 3D points across multiple frames. Instead, it learns the appropriate weights to perform motion compensation for 3D points in the feature domain. We do not employ explicit loss function for motion estimation but perform end-to-end training.

The multiscale features from the previous frame, $P^1_{0ds}, P^1_{1ds}, P^1_{2ds}, P^1_{3ds}$, and the three downsampled coordinates from the current frame, $C^2_{3ds}$, are fed to the prediction network to obtain current frame's predicted three-times downsampled tensor $\widehat{P^2_{3ds}} = \{C^2_{3ds}, \widehat{F^2_{3ds}}\}$. The prediction network downscales the input three times while concatenating it with the corresponding scale features. Finally a version of Generalized Sparse Convolution (GSC) is employed to map features from $C^1_{3ds}$ to $C^2_{3ds}$ obtain the tensor $\widehat{P^2_{3ds}}$. 

GSC is defined in \cite{minkowski} as a generalized version of sparse convolution that incorporates all discrete convolutions as special cases. Let $x^{\text{in}}_u\in\mathbb{R}^{N^{\text{in}}}$ be an $N^{\text{in}}$-dimensional input feature vector in a $D$-dimensional space at $u \in \mathbb{R}^D$ (a $D$-dimensional coordinate), and convolution kernel weights be $W \in \mathbb{R}^{K^D \times N^{\text{out}} \times N^{\text{in}}}$. The conventional dense convolution in $D$-dimension is defined in \cite{minkowski} as:
\begin{equation}
x^{\text{out}}_{u} = \sum\limits_{i\in \mathcal{V}^D(K)} W_i x^{\text{in}}_{u+i} \text{ for } u \in \mathbb{Z}^D
\label{dense_convolution}
\end{equation}
where $\mathcal{V}^D(K)$ is the list of offsets in $D$-dimensional hypercube centered at the origin with kernel size $K$. The generalized sparse convolution is defined in \cite{minkowski} as:
\begin{equation}
x^{\text{out}}_{u} = \sum\limits_{i\in \mathcal{N}^D(u,C^{\text{in}})} W_i x^{\text{in}}_{u+i} \text{ for } u \in C^{\text{out}}
\label{GSC}
\end{equation}
where $\mathcal{N}^D$ is a set of offsets that define the shape of a kernel and $\mathcal{N}^D(u,C^{\text{in}}) = \{i|u + i \in C^{in}, i \in \mathcal{N}^D\}$ as the set of offsets from the current center, $u$, that exist in $C^{in}$. $C^{in}$ and $C^{out}$ are predefined input and output coordinates of sparse tensors. In GSC, the input and output coordinates are not necessarily the same and the shape of the convolution kernel is arbitrarily defined with $\mathcal{N}^D$. 

\begin{figure}[!t]
\centering
\subfloat[\textit{Conv: Same in/out coords}]{\includegraphics[width=0.45\linewidth]{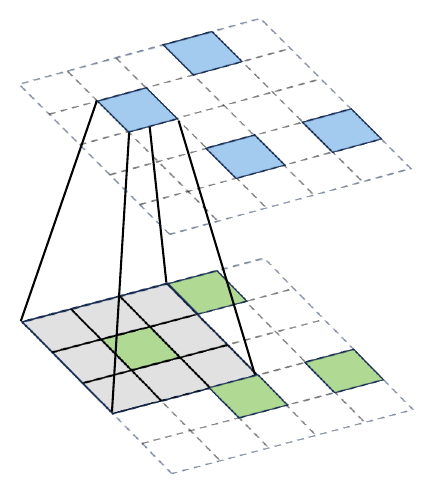}\label{sparse_conv}}
\hfil
\subfloat[\textit{GSConv: Arbitrary in/out coords}]{\includegraphics[width=0.45\linewidth]{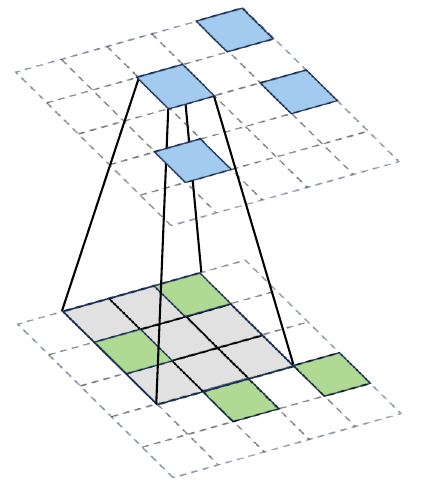}
\label{gsc}}
\caption{Comparison between the two generalized sparse convolutions employed in the proposed framework. Shown in 2D with blue as the output coordinates ($C^{\text{out}}$) and green as the input coordinates ($C^{\text{in}}$).}
\label{convolutions}
\vspace{-2mm}
\end{figure}

The proposed framework employs convolution kernel shape ($\mathcal{N}^D$) of a 3x3x3 cuboid. 
The framework employs two kinds of generalized sparse convolutions which are shown in Fig. \ref{convolutions}. 
Sparse convolutions denoted by \textit{Conv} has the same input ($C^{\text{in}}$) and output ($C^{\text{out}}$) coordinates as shown in Fig. \ref{sparse_conv} and is employed in encoder, decoder and predictor networks.
Sparse convolution denoted by \textit{GSConv} has different input ($C^{\text{in}}$) and output ($C^{\text{out}}$) coordinates as shown in Fig. \ref{gsc} and is employed only once in the predictor network.

\textit{GSConv} is employed towards the end of the predictor network to map the features from the input coordinates $C^1_{3ds}$ to the output coordinates $C^2_{3ds}$ while applying a convolution operation.
\textit{GSConv} performs motion estimation in the feature domain by translating the latent features from the downsampled coordinates of $P^1$, i.e., $C^1_{3ds}$ to the downsampled coordinates of $P^2$, i.e., $C^2_{3ds}$. This way the \textit{GSConv} enables us to predict learned features for the current frame coordinates using the previous frames multi-scale coordinates.

\subsection{Training}
During training, we optimize the Lagrangian loss, i.e., 
\begin{eqnarray}
J_{loss} = R + \lambda D
\end{eqnarray}
Where $R$ is the compressed bit rate and $D$ is the distortion loss. We employ three binary cross-entropy losses at three different scales such that the total distortion loss is: 
\begin{eqnarray}
D = \mathcal{L}_1(\widetilde{P}_{2ds},P_{2ds}) + \mathcal{L}_2(\widetilde{P}_{1ds},P_{1ds}) + \mathcal{L}_3(\widetilde{P},P)
\end{eqnarray}
Where the ground-truth $P_{2ds}$ and $P_{1ds}$ are obtained by voxel or quantization-based downsampling of the original point cloud $P$.

The three downsampled coordinates $C^2_{3ds}$ are transmitted losslessly using Octree encoder in G-PCC \cite{gpcc} and consumes a very small amount of bits (i.e., around 0.024 bpp for 8iVFB dataset \cite{8ijpeg}). We subtract the three downsampled predicted features $\widehat{F^2_{3ds}}$ from the original three downsampled features $F^2_{3ds}$ to obtain the residual features $R^2_{3ds}$. 

The residual features $R^2_{3ds}$ are quantized before encoding. Note that the quantization operation is non-differentiable, thus during training, we approximate the quantization process by adding a uniform noise $\mu \sim \mathcal{U}(-0.5, 0.5)$. Quantized residual features (lets call them $\bar{f}$) are encoded by an arithmetic encoder using a fully factorized probabilistic entropy model \cite{balle2018variational} to estimate the probability distribution of $\bar{f}$, i.e., $p_{\bar{f}|\phi} (\bar{f}|\phi)$, where $\phi$ are the learnable parameters. Then the bpp of encoding $\bar{f}$ is approximated as:
\begin{equation}
\mathcal{R} = \frac{1}{N} \sum_i \log_2 p_{\bar{f}|\phi^{(i)}} (\bar{f}|\phi^{(i)})
\end{equation}
where $N$ is the number of points, and $i$ is the index of channels.

\begin{table*}[t]
\begin{center}
\caption{BD-Rate gains against the state-of-the-art methods using D1 distortion measurements.} \label{bdrate}
\begin{tabular}{l|c|c|c|c|c}
  \hline
   & G-PCC (octree) & G-PCC (trisoup) & PCGCv2 \cite{PCGCv2} & V-PCC intra & V-PCC inter
  \\
  \hline
  basketball & -89.15 & -60.28 & -32.45 & -60.46 & -48.82 \\
  exercise & -88.77 & -64.91 & -35.44 & -62.08 & -48.30 \\
  model & -86.25 & -56.75 & -33.69 & -61.93 & -51.80 \\
  redandblack & -88.99 & -48.11 & -28.31 & -59.33 & -55.58 \\
  soldier & -90.21 & -52.73 & -40.16 & -66.51 & -43.60 \\
  \hline
  Average & -88.77 & -56.69 & -34.08 & -62.69 & -52.44 \\
  \hline
\end{tabular}
\end{center}
\vspace{-2mm}
\end{table*}


\begin{table}[t]
\begin{center}
\caption{Reported Environment/Framework Variables.} \label{env}
\begin{tabular}{l|c}
  \hline
Parameter & Value \\
  \hline
GPU Type & RTX 3090 Ti \\
CPU Type & 11th Gen Intel® Core™ i9-11900F \\
Framework & Pytorch \\
Operating system & Ubuntu 20.04 LTS \\
Batch size & 5 \\
Loss functions & BCE loss \\
Learning rate policy & Adam \\
$\lambda$ values \rule{0pt}{1ex} & $\frac{1}{10}$, $\frac{1}{9}$, $\frac{1}{6}$, $\frac{1}{4}$, $\frac{1}{2.5}$, $\frac{1}{1.7}$, $\frac{1}{1.1}$ \\
No. of parameters \rule{0pt}{2.4ex} & $2{,}033{,}000$ \\
Peak Memory Usage (GPU) & 15 GB \\
  \hline
\end{tabular}
\end{center}
\vspace{-3mm}
\end{table}

\section{Experiments}
\label{sec:results}

\subsection{Experimental Setup}
For a fair comparison, we closely follow MPEG's common test conditions (CTC) \cite{CTC} and employ the same diverse datasets recommended by MPEG for deep learning-based dynamic point cloud compression. 
The performance of our framework has already been cross-checked and verified by the MPEG 3DG EE 5.3 working group experts.

\subsubsection{\textbf{Training Dataset}}
We train the proposed model using three sequences \textit{longdress, loot}, and \textit{queen}. Sequences \textit{longdress} and \textit{loot} are from 8i Voxelized Full Bodies dataset (8iVFB v2) \cite{8ijpeg}, while sequence \textit{queen} is from Technicolor (https://www.technicolor.com/fr). Each sequence has 300 frames with a frame rate of 30 fps. Each sequence has a 10-bit precision with around $800{,}000$ to $1{,}000{,}000$ points per point cloud frame. To decrease computational complexity during training, we divide the PC frames into smaller chunks by applying the same kd-tree partition on two consecutive frames.

\subsubsection{\textbf{Evaluation Dataset}}
We evaluate the performance of the proposed framework on five sequences: \textit{redandblack, soldier, basketball, exercise}, and \textit{model}. Sequences \textit{redandblack} and \textit{soldier} are from 8i Voxelized Full Bodies dataset (8iVFB v2) \cite{8ijpeg}, while sequences \textit{basketball, exercise}, and \textit{model} are from Owlii Dynamic Human Textured Mesh Sequence Dataset \cite{owlii}. Each sequence has a frame rate of 30 fps. The 10-bit precision datasets were employed in our experiments. Since deep learning-based inter-frame compression schemes process multiple frames at a time and hence have limited GPU memory, it is advised to use a maximum of 10-bit precision point clouds.

\subsubsection{\textbf{Training Strategy}}
We train our network with $\lambda = \frac{1}{10}, \frac{1}{9}, \frac{1}{6}, \frac{1}{4}, \frac{1}{2.5}, \frac{1}{1.5}, 1$. The Adam optimizer is utilized with a learning rate decayed from $0.0008$ to $0.00001$. We train the model for around $40{,}000$ batches with a batch size of $5$. We conduct all the experiments on a GeForce RTX 3090 GPU with 24GB memory.

\subsubsection{\textbf{Evaluation Metric}}
The bit rate is evaluated using bits per point (bpp), and the distortion is evaluated using point-to-point geometry (D1) Peak Signal-to-Noise Ratio (PSNR), and point-to-plane geometry (D2) PSNR. For some point clouds, the normals are not available which are required for D2-PSNR calculation. We employ Open3D's normal estimation using $20$ neighboring points by employing covariance analysis. The geometry PSNRs are obtained using MPEG's $pc\_error$ tool \cite{pcError}.
The peak value $p$ is set as 1023 for all the datasets. We plot rate-distortion curves and calculate the BD-Rate (Bj{\o}ntegaard Delta Rate) \cite{bjontegaard2001calculation} gains using D1-PSNR over different methods.

\begin{figure*}[!t]
     \centering
     \subfloat[]{\includegraphics[width=0.33\textwidth]{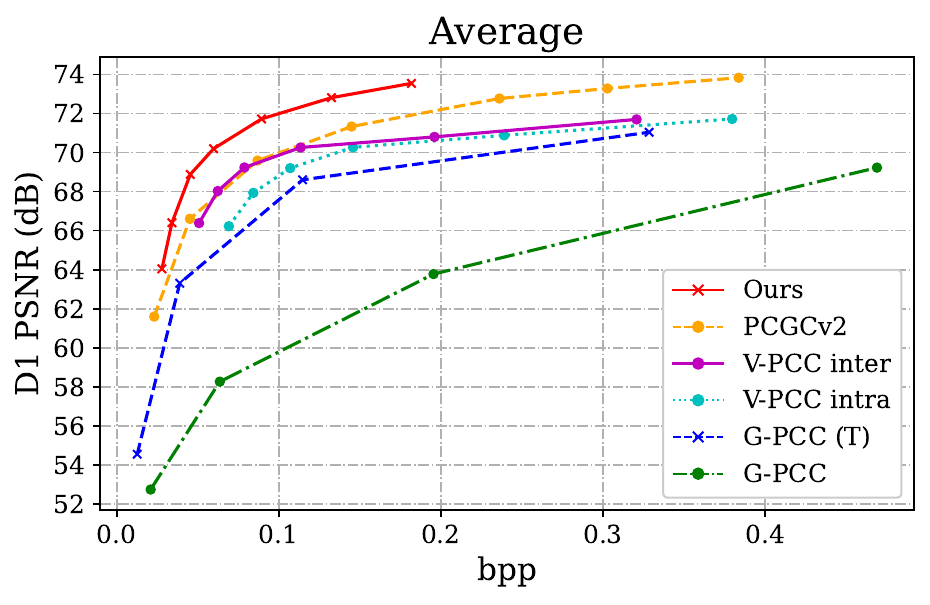}}
     \subfloat[]{\includegraphics[width=0.33\textwidth]{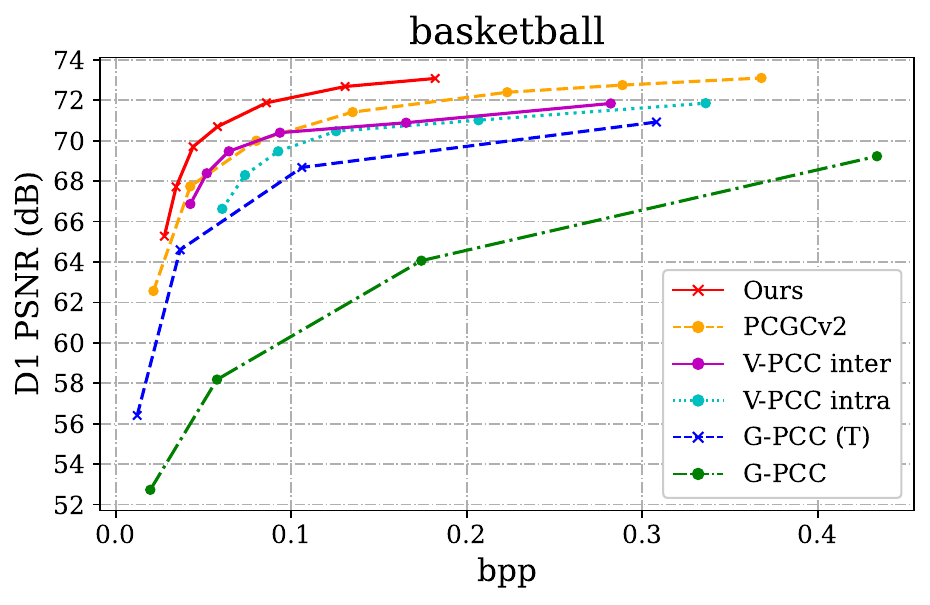}}
     \subfloat[]{\includegraphics[width=0.33\textwidth]{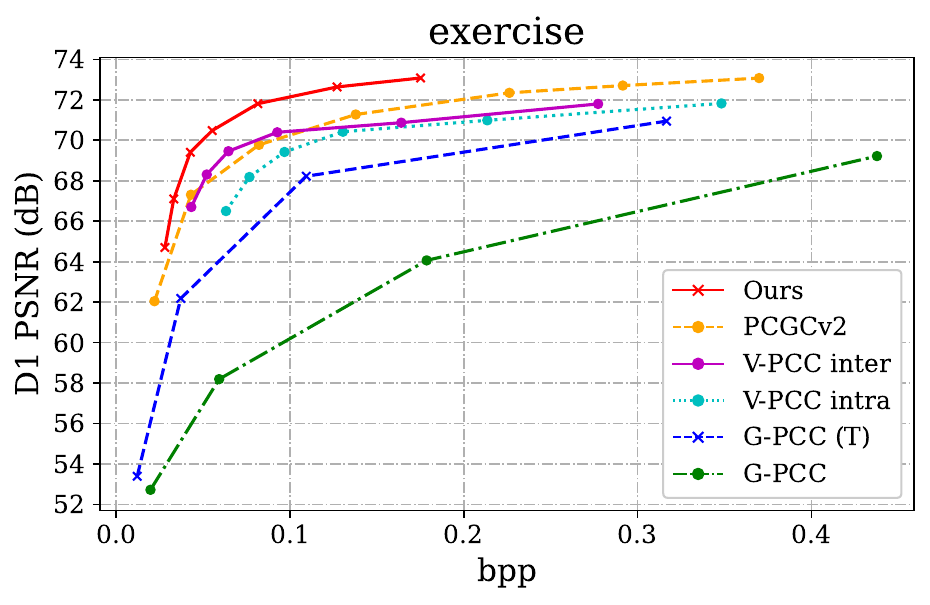}}
     \\ \vspace{-1mm}
     \subfloat[]{\includegraphics[width=0.33\textwidth]{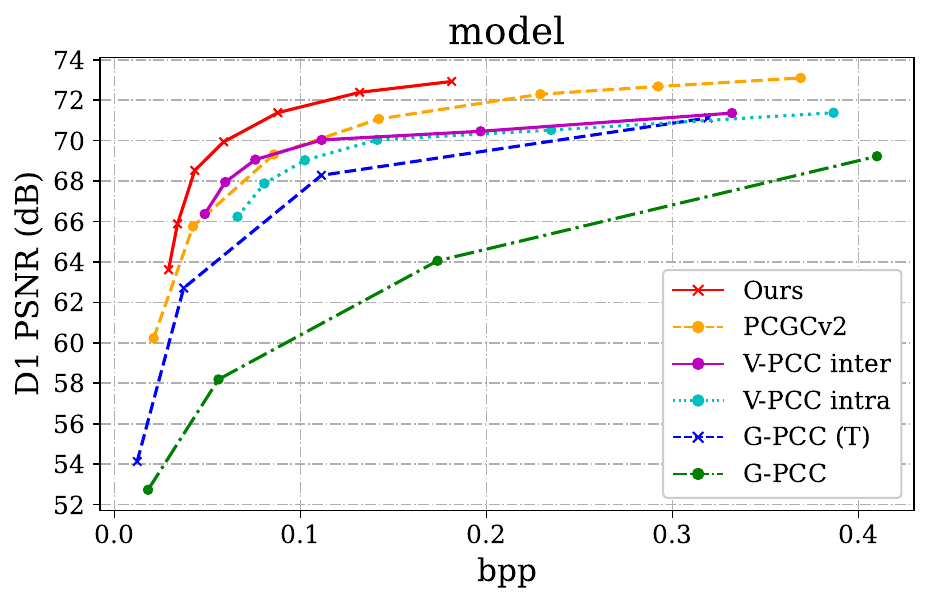}}
     \subfloat[]{\includegraphics[width=0.33\textwidth]{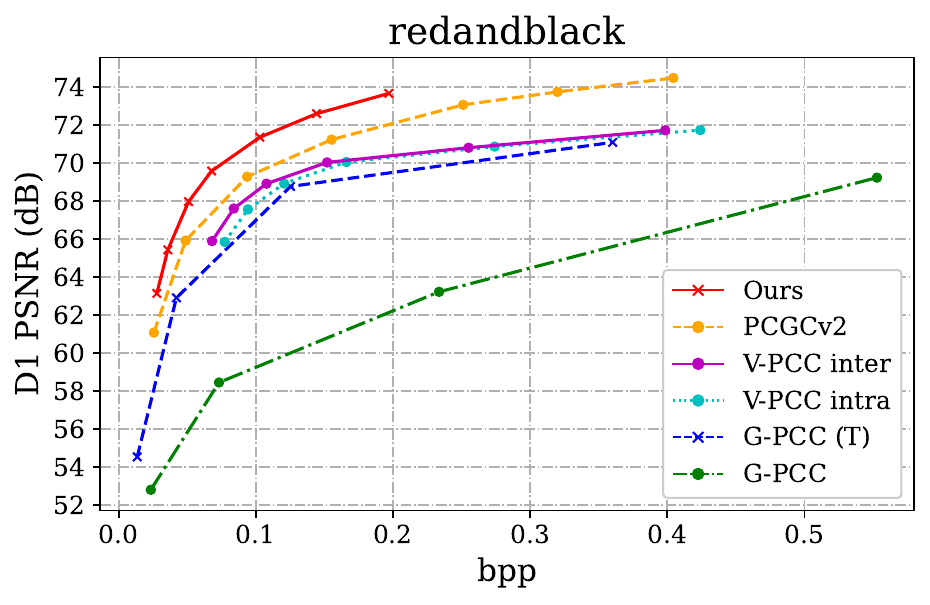}}
     \subfloat[]{\includegraphics[width=0.33\textwidth]{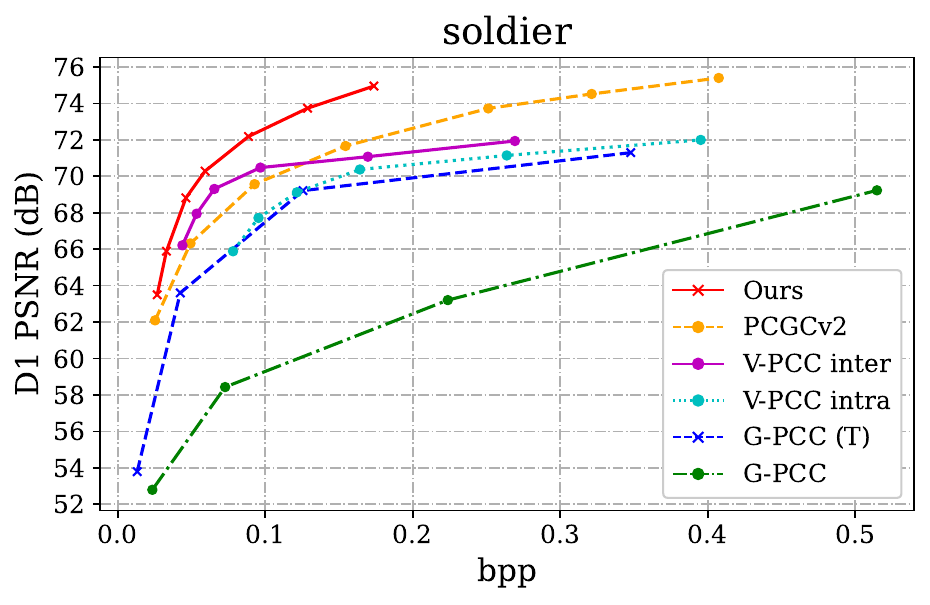}}
     \caption{Rate-distortion curves using D1 PSNR comparison with the state-of-the-arts plotted for five different sequences and their average.}
     \label{results_d1}
     \vspace{-1mm}
\end{figure*}

%

\subsection{Experimental Results}
Our framework and environment variables are shown in Table \ref{env}.
\subsubsection{\textbf{GoP Structure}}
In video coding, a group of pictures, or GOP structure, specifies the order in which intra- and inter-frames are arranged. In the experiments for our framework, the intra-frame (I frame) is encoded using PCGCv2 \cite{PCGCv2} and the inter-frame (P frame) is encoded using the proposed framework. In the results, the I frame is encoded after 32 frames and the rest of the 31 frames are encoded as inter-frames P frame. 

\subsubsection{\textbf{Baseline Setup}}
We compare our method to the state-of-the-art deep learning intra-frame encoding PCGCv2 \cite{PCGCv2}, MPEG's G-PCC (octree as well as trisoup) \cite{gpcc} methods, as well as MPEG's video-based V-PCC method (inter and intra-frame encoding) \cite{vpcc}. We utilize G-PCC's latest reference implementation TMC13-v20, and for V-PCC the latest implementation TMC2-v18 which uses the HEVC video codec. 
V-PCC inter-frame low-delay setting which involves P-frame encoding is employed for a fair comparison to the proposed P-frame encoding method. Two extra points for higher bpp have been added for V-PCC.

\begin{figure*}[!t]
     \centering
     \subfloat[]{\includegraphics[width=0.33\textwidth]{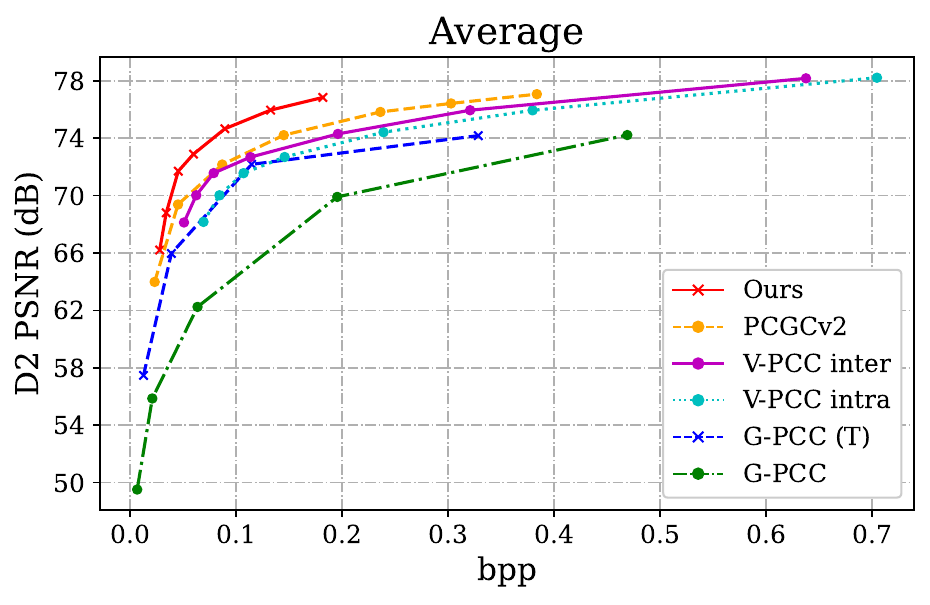}}
     \subfloat[]{\includegraphics[width=0.33\textwidth]{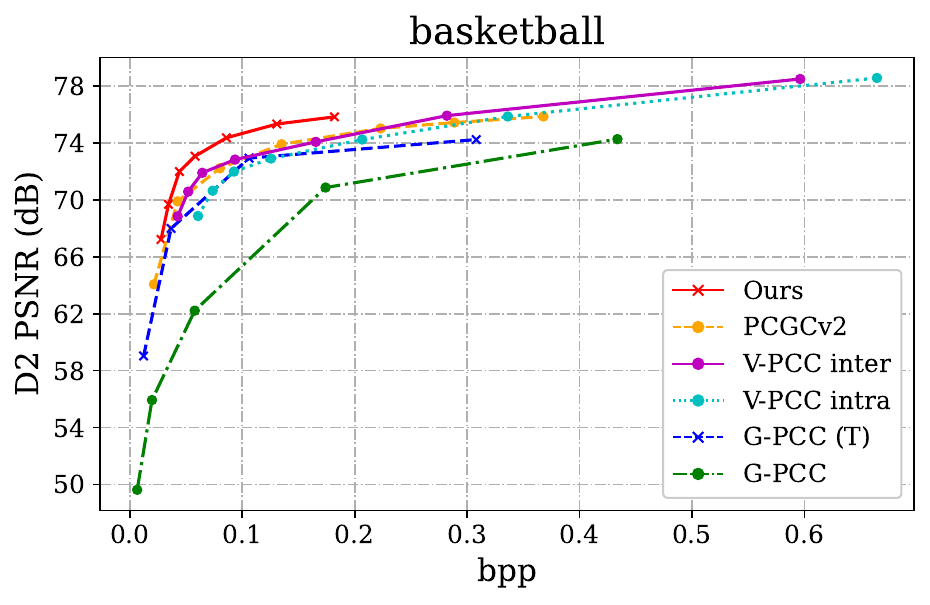}}
     \subfloat[]{\includegraphics[width=0.33\textwidth]{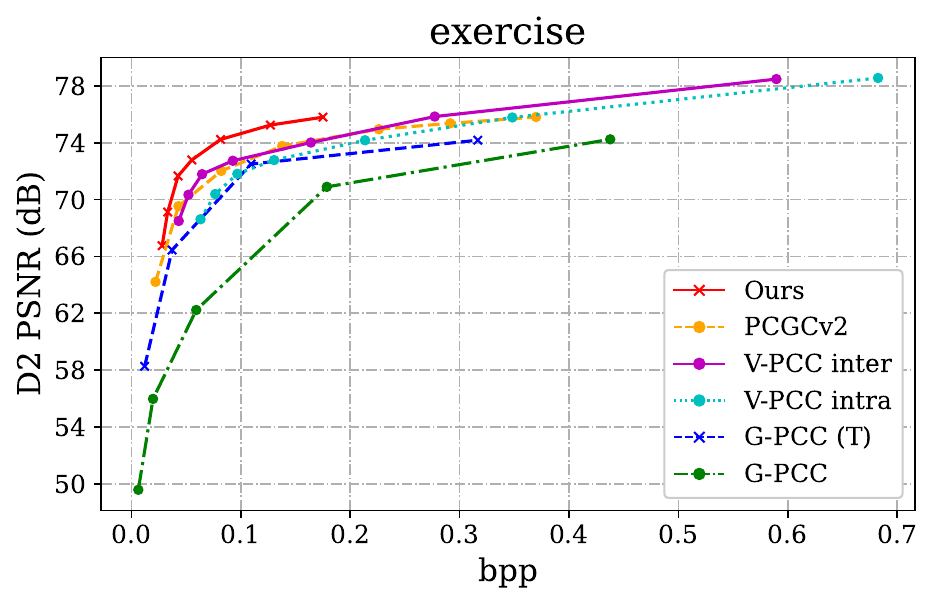}}
     \\ \vspace{-1mm}
     \subfloat[]{\includegraphics[width=0.33\textwidth]{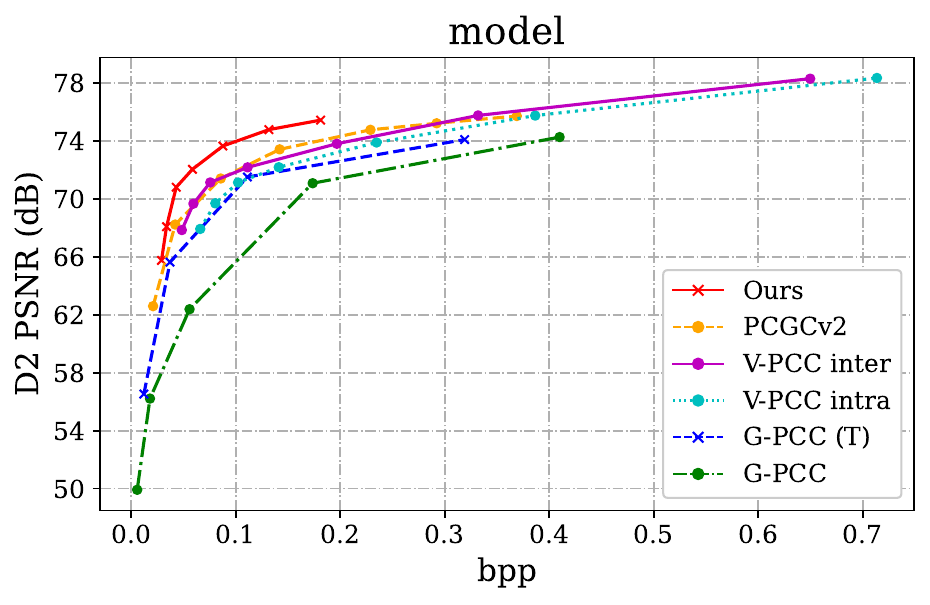}}
     \subfloat[]{\includegraphics[width=0.33\textwidth]{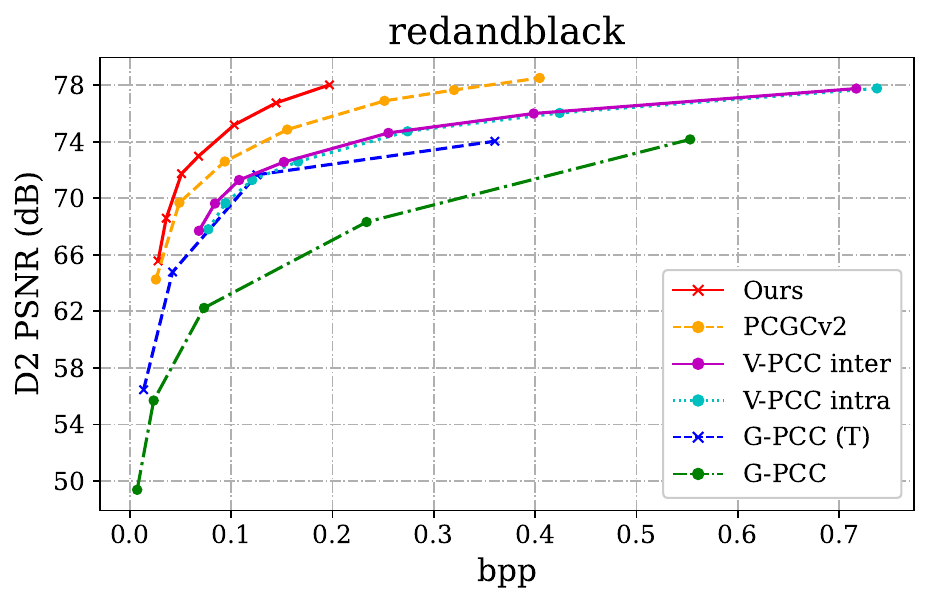}}
     \subfloat[]{\includegraphics[width=0.33\textwidth]{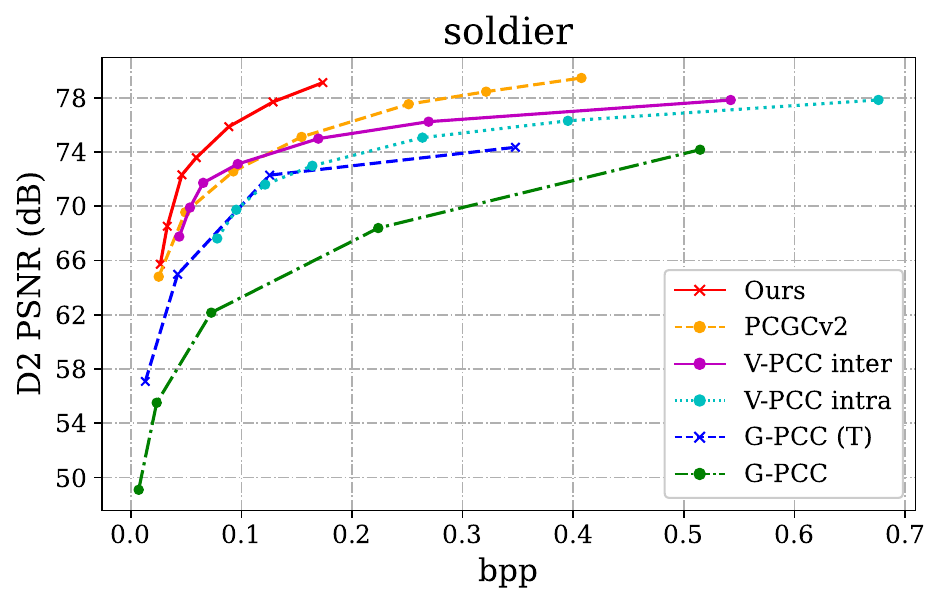}}
     \caption{Rate-distortion curves using D2 PSNR comparison with the state-of-the-arts plotted for five different sequences and their average.}
     \label{results_d2}
     \vspace{-1mm}
\end{figure*}



\begin{figure}
\centering
\includegraphics[width=\linewidth]{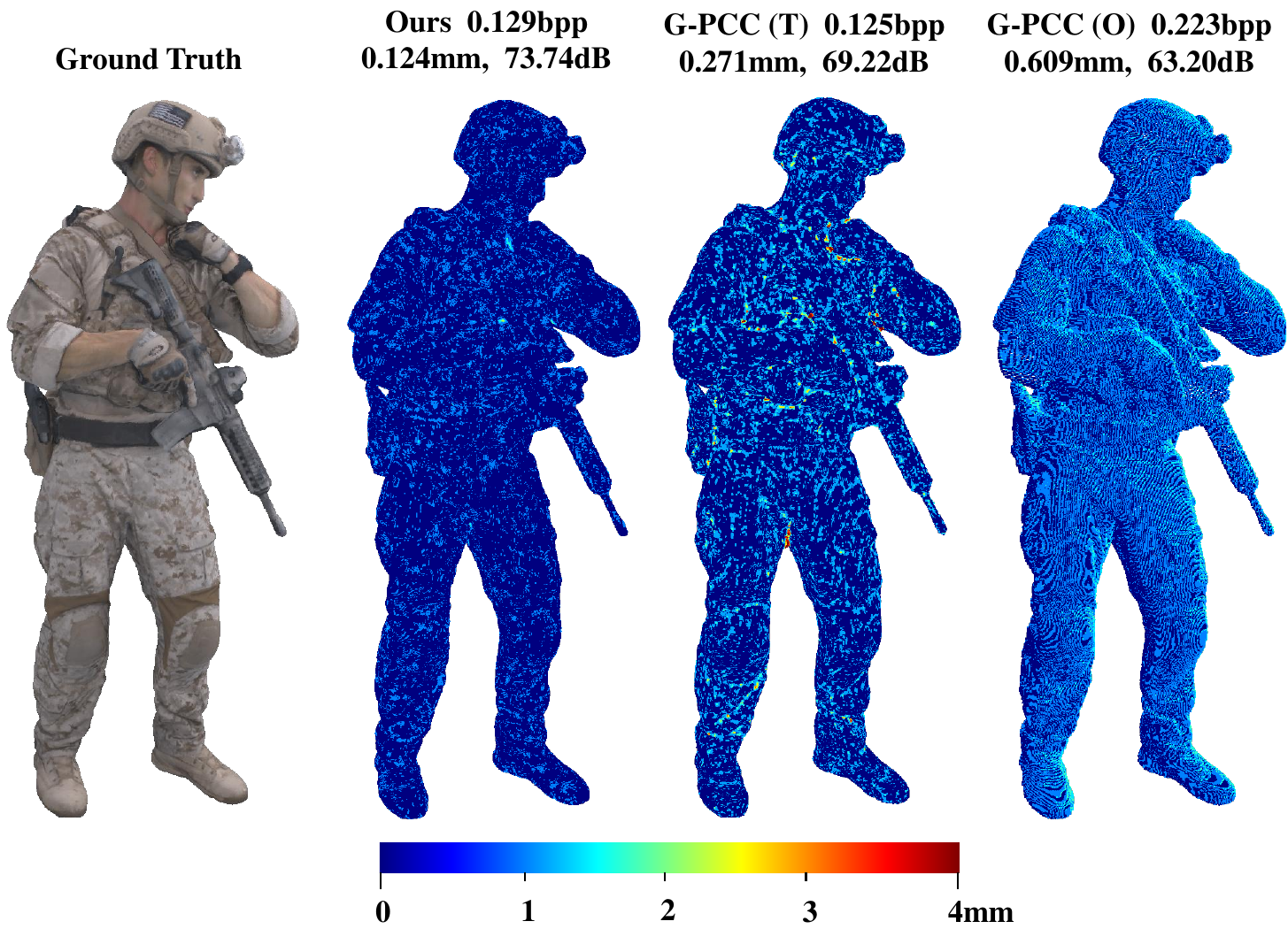}
\caption{Qualitative visual comparison of sequence ``soldier" for different methods. The color error map describes the point-to-point distortion measured in mm, and the numbers above represent the bitrate, mean error measured in mm, and D1 PSNR.}
\label{vis_res}
\vspace{-1mm}
\end{figure}

\subsubsection{\textbf{Performance Evaluation}}
Table \ref{bdrate} shows the BD-Rate gains of the proposed method over the state-of-the-art using D1-PSNR. The lower the BD-Rate value, the more the improvement is. Our method achieves significant gains compared to G-PCC with an average of $88.77\%$ BD-Rate gains against G-PCC (octree), $56.69\%$ BD-Rate improvement over G-PCC (trisoup). Compared to the deep learning-based model PCGCv2, we achieve a $34.08\%$ BD-Rate improvement. Compared to the V-PCC, we achieve a $62.69\%$ BD-Rate improvement over intra-frame encoding mode and $52.44\%$ BD-Rate improvement over inter-frame encoding mode. The proposed method outperforms V-PCC inter-frame mode across all rates for dense photo-realistic point clouds. Please note that in a previous version of this publication, we showed a $91.68\%$ BD-Rate gains against G-PCC (octree) and $84.41\%$ BD-Rate improvement over G-PCC (trisoup) but those gains were against TMC13-v14 but now we have updated G-PCC results to TMC13-v20.

The D1-PSNR and D2-PSNR rate-distortion curves are shown in Fig. \ref{results_d1} and Fig. \ref{results_d2} respectively.
As can be seen, the proposed method has significant coding gains compared with the deep learning-based model PCGCv2. It should be noted that compared with PCGCv2, our method performs much better at higher PSNR and still performs better than PCGCv2 at lower PSNRs. This is because both the proposed method and PCGCv2 transmit the three downsampled coordinates in a lossless manner and their corresponding features in a lossy manner. However, at lower PSNRs, most of the bits are consumed by coordinates (i.e., around 0.024 bpp) which constitutes the majority of the bitrate. At higher PSNR values most of the bitrates are due to features. Our inter-frame prediction network transmits only the residual of the features and, hence, can significantly decrease the feature bits transmitted leading to much higher gains at higher PSNR and bitrates.

The proposed method also significantly outperforms G-PCC (octree) as well as G-PCC (trisoup). We can notice that G-PCC trisoup performs much better than G-PCC octree which is because trisoup performs better for denser point clouds whereas octree performs better for sparse LiDAR-based point clouds. When compared with V-PCC, we can see that the proposed method achieves a much higher PSNR for the same bitrate for all of the sequences and bitrates. As expected, the V-PCC inter-frame encoding mode performs better than V-PCC intra-frame encoding mode. The sequences that have the most movement (i.e., redandblack) the V-PCC inter and V-PCC intra modes perform pretty similarly whereas the sequence with the least amount of movement (i.e., soldier) the V-PCC  inter-frame encoding method performed much better than V-PCC intra-frame encoding method. We can see a similar pattern between our proposed inter-frame method and PCGCv2 which is an intra-frame method. We see that our proposed inter-frame method has the most improvement over PCGCv2 on the soldier sequence and the least improvement over PCGCv2 on the redandblack sequence. Our Prediction module maps the features extracted from the previous frame to the coordinates extracted from the current frame. In this way, when the motion between adjacent frames is small, the performance is significantly improved.

\subsubsection{\textbf{Visual Results}}
Visual comparison of dense point clouds for geometry only is difficult since it is difficult to differentiate the quality by viewing only the points without color/attribute. The best and most common way to visualize the reconstruction results is to view the per-point distortion error. A qualitative comparison with the proposed method and G-PCC is presented in Fig. \ref{vis_res}. Point clouds are colored with the point-to-point reconstruction error for visualization. As can be seen in the visual results too, our method has a much better reconstruction quality compared to G-PCC. Another thing to notice is that the proposed method has very few outliers and generates points very close to the surface of the original point cloud.

\subsection{Runtime Comparison}
We compare the runtime of different methods in Table \ref{runtime}. We use an Intel Core i9-11900F CPU and an Nvidia GeForce GTX 3090 GPU. G-PCC runtime is computed for the highest bitrate on a CPU. While both PCGCv2 and Our method utilize the GPU. Due to the diversity in platforms, e.g., CPU vs. GPU, Python vs. C/C++, etc, the running time comparison only serves as the intuitive reference to have a general idea about the computational complexity. As can be seen, our method experiences a slight increase in runtime due to processing two PC frames at a time. However, the increased complexity is still minimal given that our network is an inter-frame prediction scheme. PCGCv2 has about $778$ thousand parameters, whereas, the proposed method has about $2{,}033$ thousand parameters which is still a relatively small network. 
The runtime complexity can be optimized by migrating to a C++ implementation and simplifying the framework. 

\begin{table}[t]
\begin{center}
\caption{Average runtime (per frame) of different methods using 8iVFBv2 PCs.} \label{runtime}
\begin{tabular}{l|c|c|c|c}
  \hline
   & G-PCC (O) & G-PCC (T) & PCGCv2 \cite{PCGCv2} & Ours
  \\
  \hline
  Enc (s) & 1.50 & 5.625 & 0.258 & 0.364  \\
  Dec (s) & 0.42 & 1.61 & 0.537 & 0.714  \\
  \hline
\end{tabular}
\end{center}
\vspace{-1mm}
\end{table}



\begin{table}[t]
\begin{center}
\caption{Partitioning the point cloud into a smaller number of blocks. Tested on soldier sequence} \label{blocks}
\begin{tabular}{c|c|c}
  \hline
  $\#$ of blocks & PSNR & bpp
  \\
  \hline
  1 & 74.56 & 0.1944  \\
  2 & 74.52 & 0.1987  \\
  4 & 74.48 & 0.2055  \\
  8 & 74.35 & 0.2158  \\
  \hline
\end{tabular}
\end{center}
\vspace{-2mm}
\end{table}

\subsection{Ablation Study: Block Size}
Even though in our evaluations, we have used the full point cloud during inference. We wanted to see the effects on PSNR and bitrate of dividing the point cloud into smaller blocks for encoding. The purpose is to demonstrate that if needed, a large point cloud can be partitioned into blocks for processing. During the encoding, we save the \textit{coordinate bitstream, feature bitstream, number of points}, and \textit{the entropy model header information} into four different files. Overall bitrate is decided by the collective size of these files. Once we divide the point cloud into blocks, each block would be encoded separately into four different files so we should expect to see a higher overhead involved. kd-tree partitioning is employed to divide each point cloud into multiple blocks and encoded the blocks independently. The results of this experiment on \textit{soldier} sequence are shown in Table \ref{blocks}. We notice that partitioning the point cloud into smaller blocks decreases the PSNR slightly. However, the difference is minimal. We also notice that the bitrate increases a bit but that could be from the overhead of saving the information in lots of files (e.g. for 8 $\#$ of blocks, we have a total of 24 files encoded, whereas, for 1 block, we have a total of 4 files encoded). It is possible to merge these files into a single file to decrease the overhead. However, that is out of the scope of the current work.

\section{Conclusion}
This work proposes a deep learning-based inter-frame compression scheme for dynamic point clouds that encodes the current frame using the decoded previous frame. We employ an encoder to obtain multi-scale features and a decoder to hierarchically reconstruct the point cloud by progressive scaling. 
The paper introduces a novel inter-prediction module that predicts the latent representation of the current frame by mapping the latent features of the previous frame to the downsampled coordinates of the current frame using a specific version of generalized sparse convolution (\textit{GSConv}) with an arbitrary input and output coordinates. The proposed method effectively performs motion estimation across frames for dynamic point clouds and encodes and transmits only the residual of the predicted features and the actual features. Sparse convolutions are employed to reduce the space and time complexity which allows the network to process two consecutive point cloud frames per inference. Exhaustive experimental results show more than $88\%$ BD-Rate gains over the state-of-the-art MPEG G-PCC (octree), more than $56\%$ BD-Rate gains over G-PCC (trisoup), more than $34\%$ BD-Rate gains over intra-frame network PCGCv2, more than $62\%$ BD-Rate improvement over MPEG V-PCC intra-frame encoding mode, and more than $52\%$ BD-Rate improvement over MPEG V-PCC inter-frame encoding mode. The proposed method has been verified in MPEG's cross-check.

\bibliographystyle{IEEEtran}
\bibliography{References}


\begin{IEEEbiography}
[{\includegraphics[width=1in,height=1.25in,clip,keepaspectratio]{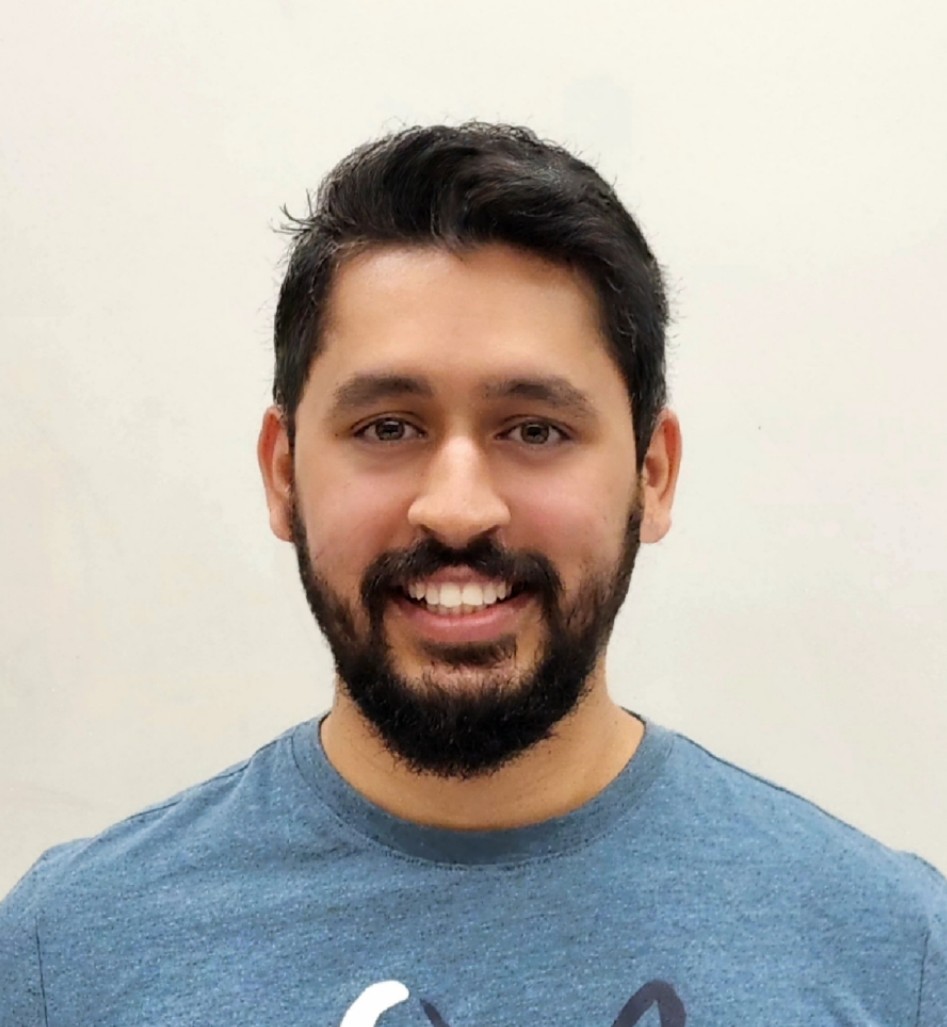}}]{Anique Akhtar} is a senior engineer at Qualcomm Technologies Inc. in San Diego, CA, USA, in the Multimedia R\&D \& Standards group where he actively participates and contributes to the standardization efforts on MPEG's Point Cloud Compression (PCC) and Video-Based Dynamic Mesh Coding (V-DMC). Anique received his B.S. degree in Electrical Engineering from Lahore University of Management Sciences (LUMS), Lahore, Pakistan, in 2013, received his M.S. degree from Koc University, Istanbul, Turkey, in 2015, and received his Ph.D. at University of Missouri-Kansas City (UMKC), USA, in 2022. He has worked in wireless communication lab as well as multimedia and communication lab in the past. His research interests include immersive video, point cloud and mesh compression, XR, and deep learning solutions for 3D data compression.\end{IEEEbiography}

\begin{IEEEbiography}
[{\includegraphics[width=1in,height=1.25in,clip,keepaspectratio]{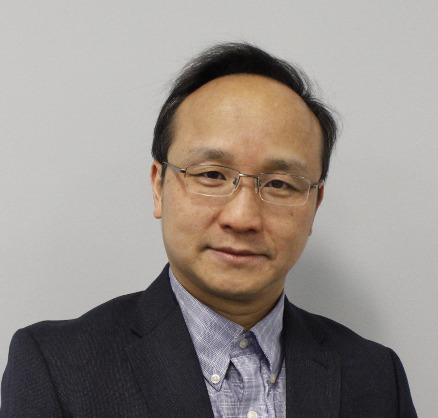}}]{Zhu Li} is a professor with the Dept of Computer Science \& Electrical Engineering, University of Missouri, Kansas City, and the director of NSF I/UCRC Center for Big Learning (CBL) at UMKC. He received his PhD in Electrical \& Computer Engineering from Northwestern University in 2004. He was AFRL summer faculty at the UAV Research Center, US Air Force Academy (USAFA), 2016-18, 2020-23. He was senior staff researcher with the Samsung Research America's Multimedia Standards Research Lab in Richardson, TX, 2012-2015, senior staff researcher with FutureWei Technology's Media Lab in Bridgewater, NJ, 2010~2012, assistant professor with the Dept of Computing, the Hong Kong Polytechnic University from 2008 to 2010, and a principal staff research engineer with the Multimedia Research Lab (MRL), Motorola Labs, from 2000 to 2008. His research interests include point cloud and light field compression, graph signal processing and deep learning in the next gen visual compression, image processing and understanding. He has 50+ issued or pending patents, 190+ publications in book chapters, journals, and conferences in these areas. He is an IEEE senior member, associate Editor-in-Chief for IEEE Trans. on Circuits \& System for Video Tech, associated editor for IEEE Trans. on Image Processing(2020~), IEEE Trans. on Multimedia (2015-18), IEEE Trans. on Circuits \& System for Video Technology(2016-19). He received the Best Paper Award at IEEE Int'l Conf on Multimedia \& Expo (ICME), Toronto, 2006, and IEEE Int'l Conf on Image Processing (ICIP), San Antonio, 2007.
\end{IEEEbiography}

\begin{IEEEbiographynophoto}
{Geert~Van~der~Auwera} received the Ph.D. degree in Electrical Engineering from Arizona State University, Tempe, AZ, USA, in 2007, and the Belgian MSEE degree from Vrije Universiteit Brussel (VUB), Brussels, Belgium, in 1997.
Presently, he is a Director at Qualcomm Technologies Inc. in San Diego, CA, USA, in the Multimedia R\&D \& Standards group where he actively contributes to the standardization efforts on MPEG's Dynamic Mesh Compression, Point Cloud Compression and previously on JVET's Versatile Video Coding (VVC). Until Jan. 2011, he was with Samsung Electronics in Irvine, CA, USA. Until Dec. 2004, he was Scientific Advisor with IWT-Flanders, the Institute for the Promotion of Innovation by Science and Technology in Flanders, Belgium. In 2000, he joined IWT-Flanders after researching wavelet video coding at IMEC’s Electronics and Information Processing Department (VUB-ETRO) in Brussels, Belgium. In 1998, his MSEE thesis on motion estimation in the wavelet domain received the Barco and IBM prizes by the Fund for Scientific Research of Flanders, Belgium. His research interests are point cloud compression, XR, video coding, video traffic and quality characterization, video streaming mechanisms and protocols.
\end{IEEEbiographynophoto}

\end{document}